\documentclass[lettersize,journal]{IEEEtran}
\usepackage{amsmath,amsfonts}
\usepackage{algorithmic}
\usepackage{algorithm}
\usepackage{array}
\usepackage[caption=false,font=normalsize,labelfont=sf,textfont=sf]{subfig}
\usepackage{textcomp}
\usepackage{stfloats}
\usepackage{url}
\usepackage{verbatim}
\usepackage{graphicx}
\usepackage{tabularx, booktabs}
\usepackage{cite}
\usepackage[table]{xcolor}
\usepackage{multirow}
\usepackage[hidelinks]{hyperref}
\usepackage{makecell}
\begin{document}

\title{Brain-Inspired Capture: Evidence-Driven Neuromimetic Perceptual Simulation for Visual Decoding}

\author{Feixue Shao, Guangze Shi, Xueyu Liu,~\IEEEmembership{Member,~IEEE}, Yongfei Wu,~\IEEEmembership{Member,~IEEE}, Mingqiang Wei,~\IEEEmembership{Senior Member,~IEEE}, Jianan Zhang, Jianbo Lu, Guiying Yan, and Weihua Yang
\thanks{This work was supported in part by the National Natural Science Foundation of China under Grant No. 12371356 and No. 62572339, in part by the Shanxi Key Laboratory of Digital Design and Manufacturing under Grant No. 202204010931025, in part by Open Foundation of the State Key Laboratory of Mathematical Sciences under Grant No. SKLMS-2025-KFKT-TD-03 and in part by Capital’s Funds for Health Improvement and Research under Grant No. 2026-1-4072. (Corresponding authors: Xueyu Liu and Weihua Yang)}

\thanks{Feixue Shao, Jianan Zhang and Weihua Yang are with the School of Mathematics, Taiyuan University of Technology, Shanxi 030024, China. (e-mail: 2024310264@link.tyut.edu.cn; zhangjianan@tyut.edu.cn; yangweihua@tyut.edu.cn).}
\thanks{Guangze Shi, Xueyu Liu, Yongfei Wu and Mingqiang Wei are with the College of Artificial Intelligence, Taiyuan University of Technology, Shanxi 030024, China. (e-mail: shiguangze0117@link.tyut.edu.cn; liuxueyu@tyut.edu.cn; wuyongfei@tyut.edu.cn; weimingqiang@tyut.edu.cn).}
\thanks{Jianbo Lu is with the National Human Genetics Resource Center, National Research Institute for Family Planning, Beijing 100000, China. (e-mail: jblu@lsec.cc.ac.cn).}
\thanks{Guiying Yan is with the Academy of Mathematics and Systems Science, Chinese Academy of Sciences, Beijing 100000, China. (e-mail: yangy@amt.ac.cn).}

\thanks{This work has been submitted to the IEEE for possible publication. Copyright may be transferred without notice, after which this version may no longer be accessible.}}
\markboth{IEEE TRANSACTIONS ON MULTIMEDIA,~2026}%
{Shell \MakeLowercase{\textit{et al.}}: A Sample Article Using IEEEtran.cls for IEEE Journals}



\maketitle

\begin{abstract}
Visual decoding of neurophysiological signals is a critical challenge for brain-computer interfaces (BCIs) and computational neuroscience. However, current approaches are often constrained by the systematic and stochastic gaps between neural and visual modalities, largely neglecting the intrinsic computational mechanisms of the Human Visual System (HVS). To address this, we propose Brain-Inspired Capture (BI-Cap), a neuromimetic perceptual simulation paradigm that aligns these modalities by emulating HVS processing. Specifically, we construct a neuromimetic pipeline comprising four biologically plausible dynamic and static transformations, coupled with Mutual Information (MI)-guided dynamic blur regulation to simulate adaptive visual processing. Furthermore, to mitigate the inherent non-stationarity of neural activity, we introduce an evidence-driven latent space representation. This formulation explicitly models uncertainty, thereby ensuring robust neural embeddings. Extensive evaluations on zero-shot brain-to-image retrieval across two public benchmarks demonstrate that BI-Cap substantially outperforms state-of-the-art methods, achieving relative gains of 9.2\% and 8.0\%, respectively. We have released the source code on GitHub through the link \url{https://github.com/flysnow1024/BI-Cap}.
\end{abstract}


\section{Introduction}
\IEEEPARstart{E}{lectroencephalography} (EEG)~\cite{cohen2017does}, magnetoencephalography (MEG)~\cite{da2013eeg}, and other neurophysiological signals provide direct measurements of brain activity~\cite{heeger2002does,9105088}, encoding rich information on visual perception and cognition~\cite{11433075}. Visual decoding aims to reconstruct or retrieve human-perceived visual content from these high-dimensional, non-stationary signals. This technology offers a pivotal window into the complex cognitive mechanisms of the Human Visual System (HVS) and lays a solid foundation for efficient Brain--Computer Interfaces (BCIs) and human--machine symbiotic intelligence~\cite{6046230,10636811}.

\begin{figure}
\centerline{\includegraphics[width=1\columnwidth]{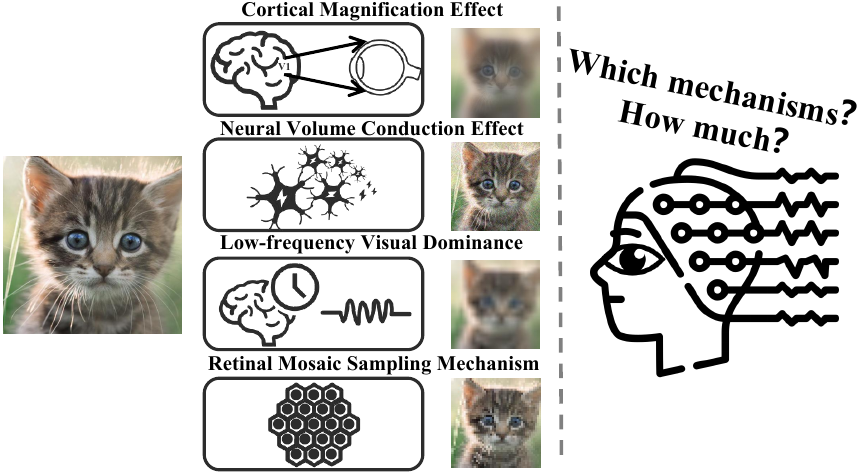}}
\caption{Which neural mechanisms should be emphasized, and to what extent, when applying biologically plausible transformations?} \label{fig1}
\end{figure}

\begin{figure*}
\centerline{\includegraphics[width=1\textwidth]{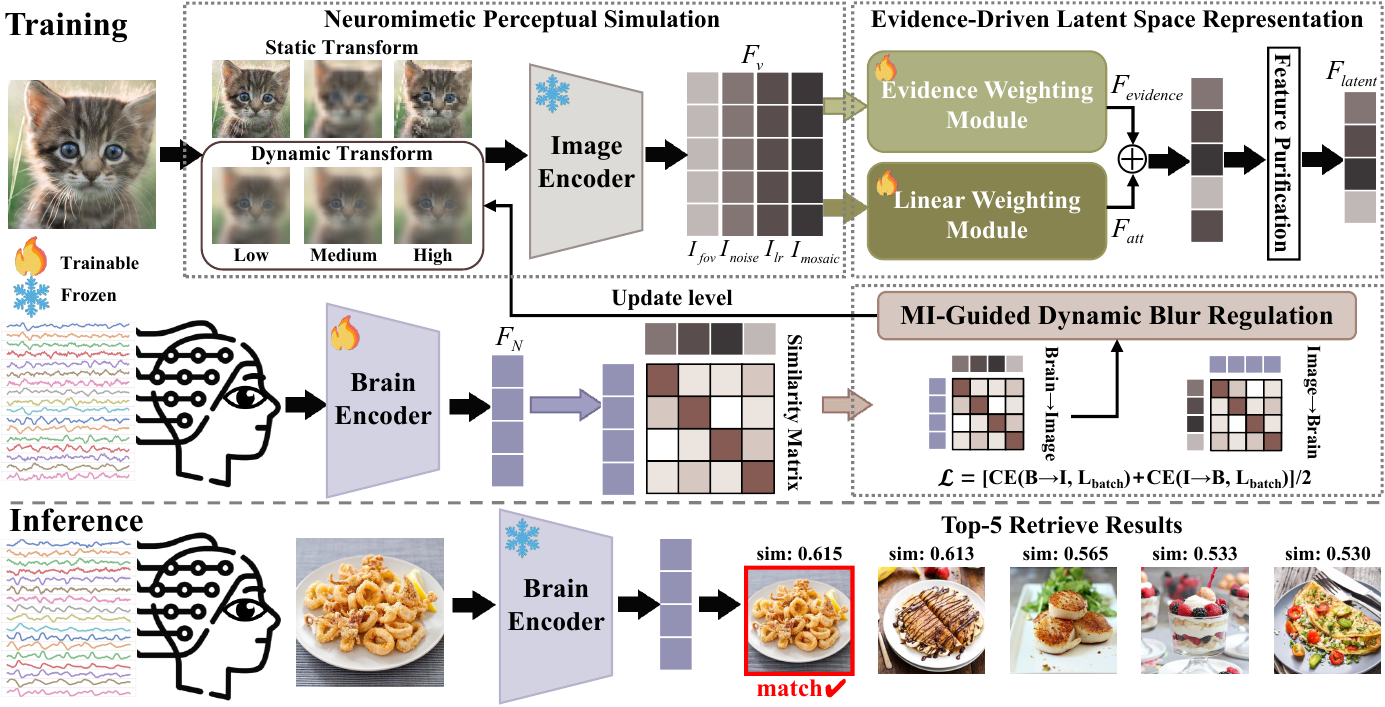}}
\caption{Overview of the proposed BI-Cap framework for EEG/MEG-based zero-shot brain-to-image retrieval. This architecture extracts visual features through neuromimetic perceptual simulation and evidence-driven latent space representation, aligns them with neural signal features, and uses MI-guided dynamic blur regulation adaptive adjustment to further optimize cross modal retrieval by adjusting the degree of foveated blur.} \label{fig2}
\end{figure*}

In recent years, with the rapid development of deep learning, significant progress has been made in visual alignment methods for brain-to-image retrieval based on neural signals. Early explorations primarily focused on identifying linear mappings between neural and visual features. Subsequently, pioneering work such as Uncertainty-aware Blur Prior (UBP)~\cite{wu2025bridging} introduced physiological mechanism priors for the first time, by simulating visual focal points through dynamic foveated Gaussian blur, it demonstrated the effectiveness of processing image paradigms based on biological mechanisms. However, UBP overlooks the intrinsic asymmetry between visual and neural modalities, which constrains the overall alignment performance.
To mitigate this, an Adaptive Teaching System (ATS) is proposed~\cite{wu2025shrinking} based on UBP, enabling the ``teacher'' (visual) modality to adjust its representations under task guidance to better match neural signals. However, relying on a single simulation mechanism fails to accurately bridge the \textbf{systematic gap} across modalities. More recently, state-of-the-art performance is achieved by applying asymmetric, modality-specific transformations to EEG signals and images~\cite{zhang2025neurobridge}. Despite these advances, such approach largely depend on linear projections and fail to capture complex non-linear mappings, while also neglecting the \textbf{stochastic gap} caused by the high dynamicity and inter-subject heterogeneity of neural activity. As a result, the intrinsic uncertainty of neural responses remains insufficiently modeled.

To address these challenges, we propose a Neuromimetic Perceptual Simulation (NPS) paradigm, which advocates achieving intrinsic cross-modal alignment through systematic simulation of HVS processing mechanisms. As illustrated in Fig.~\ref{fig1}, we design four biologically plausible transformations to simulate the cortical magnification effect~\cite{rovamo1978cortical}, the neural volume conduction effect~\cite{faisal2008noise}, the low-frequency dominance~\cite{sugase1999global,vuilleumier2003distinct} in Rapid Serial Visual Presentation (RSVP)~\cite{grootswagers2019representational}, and the retinal photoreceptor mosaic sampling mechanism~\cite{roorda1999arrangement,ramachandran1991perceptual}. These transformations construct visual representations that are physiologically highly congruent with neural signals to mitigate the systematic gap. Furthermore, to tackle the dynamicity and heterogeneity of neural activity and resolve the inherent stochastic gap, we propose evidence-driven latent space representation. By explicitly quantifying the epistemic uncertainty within individual neural processing mechanisms, this approach achieves robust dynamic perception and adaptive calibration of neurophysiological signals. Our main contributions are summarized as follows:
\begin{itemize}
    \item We propose and validate Brain-Inspired Capture (BI-Cap), a neuromimetic perceptual simulation paradigm based on four dynamic and static biologically plausible transformations, demonstrating the reliability and physiological significance of this paradigm in brain-to-image cross-modal alignment.
    \item To address the high dynamicity and heterogeneity of neural activity, we propose an evidence-driven latent space learning framework that explicitly models uncertainty in human visual system processing. Our method integrates evidential weighting with a residual attention mechanism, and then applies a feature purification module to enforce compactness and mitigate noise.
    \item We achieve state-of-the-art performance in zero-shot brain-to-image retrieval for EEG and MEG modalities, under both intra-subject and inter-subject settings.
\end{itemize}
\section{Related Work}
\label{section2}
\subsection{Multimodal Neural Signal Visual Decoding}
Early research on visual decoding primarily relied on feature extraction from single-modality fMRI signals~\cite{allen2022massive,chen2023seeing}. Although EEG and MEG offer millisecond-level temporal resolution and superior portability~\cite{spampinato2017deep,kavasidis2017brain2image}, they are constrained by coarse spatial resolution and inherent signal noise. These limitations often lead to flattened feature representations that fail to capture the brain’s progressive refinement process from low-level to high-level features.
Following the success of vision-language pre-training models such as Contrastive Language-Image Pre-training (CLIP)~\cite{radford2021learning}, the mainstream paradigm in this field has shifted toward cross-modal contrastive learning. This approach maps neural signals into a shared latent semantic space with images or text~\cite{li2024visual,songdecoding}. Representative works include BraVL~\cite{du2023decoding}, which innovatively introduce the text modality as auxiliary supervision, leveraging image captions or category descriptions to guide the learning of neural features. HA~\cite{rajabi2025human} utilizes an image-aligned encoder to map brain signals directly into a visual feature space.

Despite these advances, systematic and stochastic gaps remain between neural signals and artificial vision features~\cite{li2020perils,ahmed2021object}. To mitigate these gaps, ViEEG~\cite{liu2025vieeg} proposes a hierarchical decoupling paradigm that decomposes images into low-level edges, foreground entities, and high-level scenes via artificial cortical anchors, forcing EEG embeddings to simulate the hierarchical processing of biological vision. UBP~\cite{wu2025bridging} takes a biological perspective by introducing a dynamically adjusted blur prior to simulate early perceptual information, effectively narrowing both systematic and stochastic gaps. Building on this, ATS~\cite{wu2025shrinking} employs a teacher-student framework to resolve the asymmetry between visual and brain modalities. Furthermore, NeuroBridge~\cite{zhang2025neurobridge} facilitates cross-modal alignment by combining modality-specific cognitive prior enhancement with shared semantic projections.

\subsection{Evidential Learning in Multimodal Data}
Evidential Learning (EL), grounded in Dempster-Shafer Theory~\cite{dempster1968generalization} and Subjective Logic~\cite{jsang2018subjective}, enables explicit modeling of belief and uncertainty by parameterizing a Dirichlet distribution over the probability simplex. Unlike traditional softmax-based neural networks that often yield overconfident point estimates, EL treats the network outputs as evidence to quantify the vacuity of knowledge~\cite{li2022trustworthy,gao2025comprehensive}. By mapping evidence to the concentration parameters of the Dirichlet distribution, EL provides a principled way to distinguish between aleatoric uncertainty and epistemic uncertainty, which is particularly vital when dealing with limited or high-noise datasets. In multimodal settings, EL is primarily applied to multimodal fusion and cross-modal retrieval. For multimodal fusion, EL-based frameworks treat each modality as an independent source of evidence, aggregating modality-specific beliefs via the Dempster’s Combination Rule to ensure robust integration of heterogeneous modalities~\cite{ma2021trustworthy,gao2023collecting}. This mechanism allows the model to dynamically suppress uninformative modalities while highlighting those with high evidential support. In retrieval scenarios, formulated as bidirectional classification problems, EL has been employed to model evidential distributions and improve robustness under noisy correspondences~\cite{li2023dcel}. Representative approaches leverage bidirectional evidence to detect mismatched pairs~\cite{qin2022deep} or explicitly model data uncertainty to calibrate retrieval predictions~\cite{li2023prototype}. By calibrating retrieval predictions through evidential distributions, these methods facilitate a more reliable and trustworthy alignment between diverse heterogeneous modalities.

\section{Methods}
\subsection{Problem Statement}
The primary objective of our proposed BI-Cap framework is to retrieve the specific ground-truth image that elicits a given neural signal from a large-scale candidate pool containing \textbf{unseen images}. An overview of the BI-Cap architecture is illustrated in Fig.~\ref{fig2}. Following the zero-shot experimental protocol established in~\cite{wu2025bridging}, we utilize a paired dataset $\mathcal{D} = \{(I, N)\}$, where the training and test classes are strictly mutually exclusive ($\mathcal{D}_{train} \cap \mathcal{D}_{test} = \emptyset$) to evaluate the model's generalization to novel semantic categories.

For the visual modality, $I \in \mathbb{R}^{C_I \times H \times W}$ represents the original image data, where $C_I, H, W$ denote the number of channels, height, and width, respectively. A pre-trained image encoder is employed to extract high-level visual features $f \in \mathbb{R}^{d_v}$. The neuromimetic visual features $F_{V}=\{ {f}_v \}_{v=1}^{V} \in \mathbb{R}^{V \times d_v}$, where $V$ denotes the number of transformation and $d_v$ represents the feature dimension. For the neural modality, $N \in \mathbb{R}^{C_N \times L}$ represents the corresponding neural signal, with $C_N$ recording channels and a temporal segment of length $L$. We adopt the brain encoder proposed in the~\cite{wu2025shrinking} to extract neural features $F_N \in \mathbb{R}^{d_n}$. The core of BI-Cap lies in mapping these heterogeneous features into a unified latent space. Specifically, $F_N$ and $F_V$ undergo a series of transformations to yield the final embeddings $F_{latent}$ and $F_N$, respectively. The retrieval task is subsequently performed by identifying the image within the candidate pool whose visual embedding demonstrates the highest semantic correspondence with the given neural representation in the shared latent space, thereby establishing a precise cross-modal alignment.

\subsection{Neuromimetic Perceptual Simulation}
Inspired by the physiological characteristics of HVS, we project the original images $I$ into a neuromimetic perceptual space using four biologically plausible transformations.

To simulate the spatial non-uniformity of the cortical magnification factor, we introduce a foveation mask $M_f \in [0, 1]$, which controls the transition from the high-acuity foveal region to the blurred periphery. The mask value at spatial location $(i, j)$ is governed by an exponential decay function:
\begin{equation}
    M_f(i, j) = \exp \left( - \gamma \cdot \frac{d(i, j)}{D} \right),
    \label{eq:foveation_mask}
\end{equation}
where $d(i, j)$ denotes the Euclidean ($L_2$-norm) distance between pixel $(i, j)$ and the designated foveation center, and $D$ represents the maximum Euclidean distance within the image plane. The parameter $\gamma$ serves as a decay coefficient.
Consequently, the final foveated visual stimulus, denoted as $I_{fov}$, is synthesized by spatially blending the original image $I$ with the biologically blurred image counterpart $I_{blur}^k$. The blurred image is generated using a Gaussian blur kernel of size $k$ and perturbation constant $c$, and the fusion is implemented through element-wise multiplication $\odot$:
\begin{equation}
    I_{fov} = M_f \odot I + (1 - M_f) \odot I_{blur}^k.
    \label{eq:foveal_blending}
\end{equation}

Secondly, addressing the inherent neural volume conduction effect, we introduce additive Gaussian white noise to the visual representation. The noisy stimulus is formulated as $I_{noise} = I + \mathcal{N}(0, \sigma^2)$, where $\mathcal{N}(0, \sigma^2)$ represents a Gaussian noise distribution with zero mean and variance $\sigma^2$.

Thirdly, inspired by the coarse-to-fine low-frequency dominance mechanism of rapid visual processing, we aim to selectively filter out high-frequency stochastic redundancies that are functionally less relevant to macroscopic EEG signals. This aligns with the biological observation that early visual evoked potentials are more sensitive to global structural contours than to fine-grained textures. To achieve this, we employ \textit{bilinear interpolation} for resampling to generate the low-resolution images $I_{low}$. Simultaneously, to emulate the discrete sampling mechanism of the retinal mosaic, we utilize \textit{nearest-neighbor interpolation}. This operation mimics the quantization of the continuous light field into discrete neural receptive fields, yielding the mosaic-like images $I_{mosaic}$. These operations can be formally expressed as:
\begin{equation}
    \begin{aligned}
        I_{low} &= \mathcal{R}_{bilinear}(I, s_{low}) \\
        I_{mosaic} &= \mathcal{R}_{nearest}(I, s_{mos}),
    \end{aligned}
    \label{eq:downsampling}
\end{equation}
where $\mathcal{R}_{bilinear}$ and $\mathcal{R}_{nearest}$ denote the resampling operators based on bilinear and nearest-neighbor interpolation, respectively, and $s_{low}, s_{mos}$ represent the corresponding down-sampling scale factors.

Finally, the four distinct biologically plausible visual images generated through these transformations are parallelly fed into the pre-trained CLIP image encoder. This yields a set of aligned neuromimetic feature embeddings $F_V$:
\begin{equation}
    F_{V} = \text{Encoder}_{\text{CLIP}}(\{ I_{fov}, I_{noise}, I_{low}, I_{mosaic} \}).
    \label{eq:feature_encoding}
\end{equation}

\begin{figure}
\centerline{\includegraphics[width=1\columnwidth]{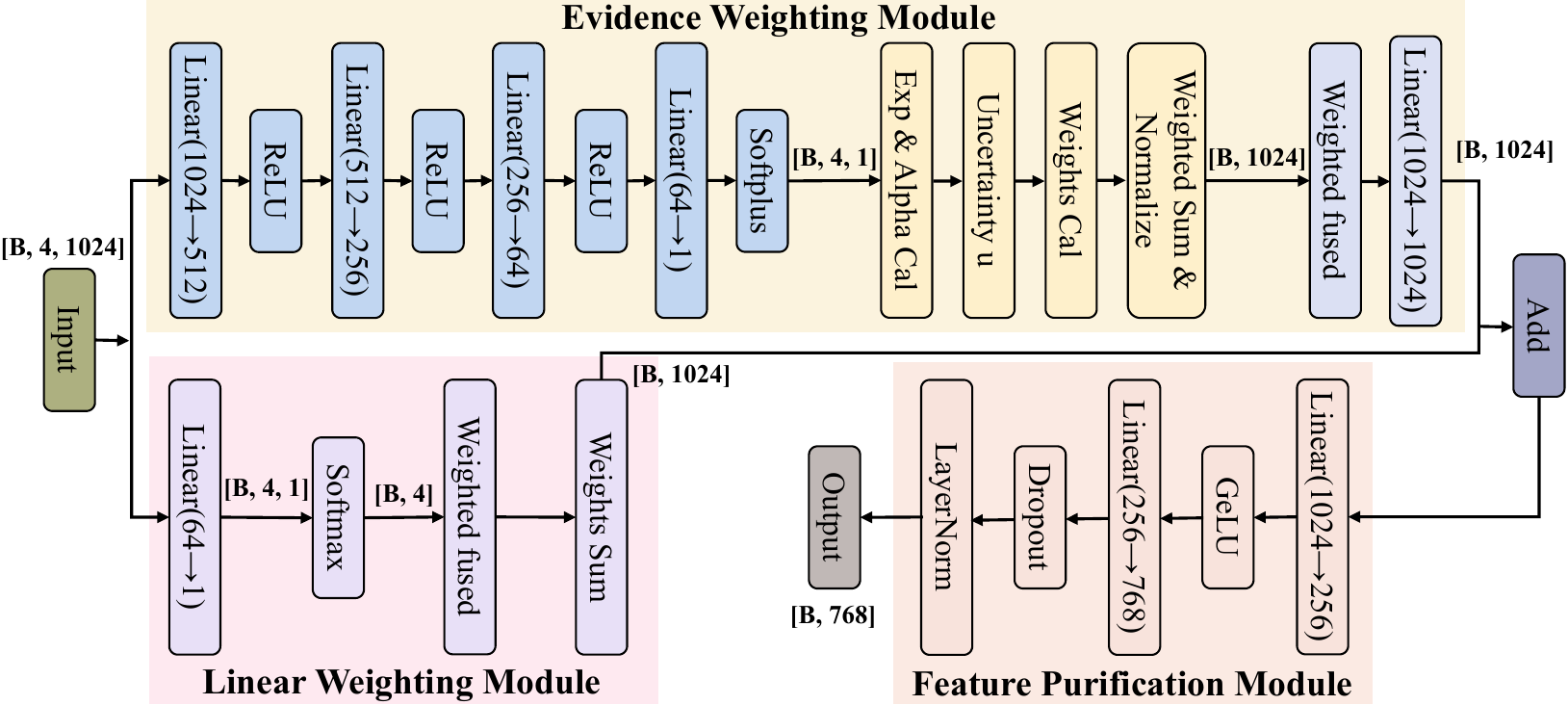}}
\caption{Overview of the proposed framework for Evidence-Driven Latent Space Representation. Features from parallel Evidence and Linear Weighting streams are integrated via element-wise addition, and then processed by a bottleneck Feature Purification Module to align visual representations.} \label{fig3}
\end{figure}

\subsection{Evidence-Driven Latent Space Representation}
\label{sec:evidence_driven}
Due to the heterogeneity and temporal variability of neural activity, different visual transformations may contribute unequally to modeling EEG responses. Therefore, as shown in Fig.~\ref{fig3}, we propose an Evidence-Driven Latent Space Representation method, which combines evidential weighting with a residual attention mechanism, then applies a feature purification module to enforce compactness and mitigate noise. These components construct a robust shared latent space. EL models belief and uncertainty associated with different hypotheses in model predictions, drawing upon Dempster-Shafer Theory and Subjective Logic. 

\subsubsection{Evidence Weighting Module}
To adaptively aggregate features from diverse neuromimetic transformations, we propose an Evidence Weighting Module (EWM) based on Subjective Logic. Unlike vanilla attention, EWM explicitly models the reliability of each transformation.
Given the input neuromimetic visual features $F_{V}$, we feed the $F_{V}$ into a non-linear Evidence Head ($h_{\theta}$). The $h_{\theta}$ is composed of a Multi-Layer Perceptron (MLP) activated by the Softplus function to ensure non-negative outputs, thereby obtaining the non-negative evidence quantity $e_v$ for each transformation:
\begin{equation}
    e_v = \exp(\text{Softplus}(\text{MLP}(F_{V}))).
    \label{eq:evidence_extraction}
\end{equation}

Following Subjective Logic, we characterize each distinct neuromimetic transformation as an independent and autonomous evidence source that provides a specific perspective on the latent neural representation. In our implementation, we focus on a single-dimensional confidence estimation to model the inherent epistemic uncertainty of each branch. Under this configuration, the Dirichlet strength $S_v$ is simplified as:
\begin{equation}
    S_v = e_v + 1.
\end{equation}
We can derive the corresponding epistemic uncertainty $u_v \in (0, 1]$ for the $v$-th transformation are derived as:
\begin{equation}
    \begin{aligned}
        u_v &= \frac{1}{S_v} = \frac{1}{e_v + 1}.
    \end{aligned}
    \label{eq:uncertainty_calculation}
\end{equation}

As the collected evidence $e_v$ increases, $S_v$ grows, and the uncertainty $u_v$ approaches 0. We define the Belief Weight $w_k$ of each transformation as its degree of certainty: $w_v = 1 - u_v$. Consequently, a higher evidence quantity $e_v$ leads to a belief weight $w_v$ approaching 1, while lack of evidence causes $w_v$ to vanish, effectively suppressing unreliable reconstructions. Utilizing these weights, we perform a weighted average on the multi-transformation features. This mechanism ensures that features with high evidence support dominate the final representation, while highly uncertain features are suppressed. The fused evidential feature $F_{evidence}$ is calculated as:
\begin{equation}
    F_{evidence} = \text{Proj}\left( \frac{\sum_{v=1}^{V} w_v \cdot f_v}{\sum_{v=1}^{V} w_v + \epsilon} \right),
    \label{eq:evidence_fusion}
\end{equation}
where $\text{Proj}(\cdot)$ is a linear projection layer.
\subsubsection{Linear Weighting Module}
To capture potential contextual dependencies and supplement global information that might be lost through pure evidential weighting, we introduce a parallel Linear Fusion Module as a residual term outside the evidence fusion path. This module computes attention scores $s_v$ via a learnable linear layer and normalizes them using the softmax function to obtain attention weights. The attention-fused feature is then computed as a weighted sum of features:
\begin{equation}
    F_{att} = \sum_{v=1}^{V} \text{softmax}(s_v) \cdot f_v.
\end{equation}
The fused features is obtained by summing the evidence-fused and the attention-fused features $F_{fus} = F_{evidence} + F_{att}$.
\subsubsection{Feature Purification Module}
To ensure feature compactness and mitigate potential noise, $F_{fus}$ is processed by the Feature Purification module with a bottleneck architecture to generate the final latent space representation $F_{latent}$. Specifically, the fused feature $F_{fus}$ is first compressed into a lower-dimensional bottleneck space via a linear layer. This process employs a GeLU activation to perform non-linear filtering, effectively suppressing task-irrelevant noise while preserving core semantic information. Subsequently, a secondary linear layer projects these refined features into a reconstructed latent space. To enhance the robustness of the cross-modal alignment, Dropout and Layer Normalization are integrated to stabilize the feature distribution and prevent overfitting. The resulting purified embedding $F_{latent}$ is the final representation for achieving cross modal alignment.

\begin{figure}[!t]
\centering
\subfloat[]{\includegraphics[width=1.75in]{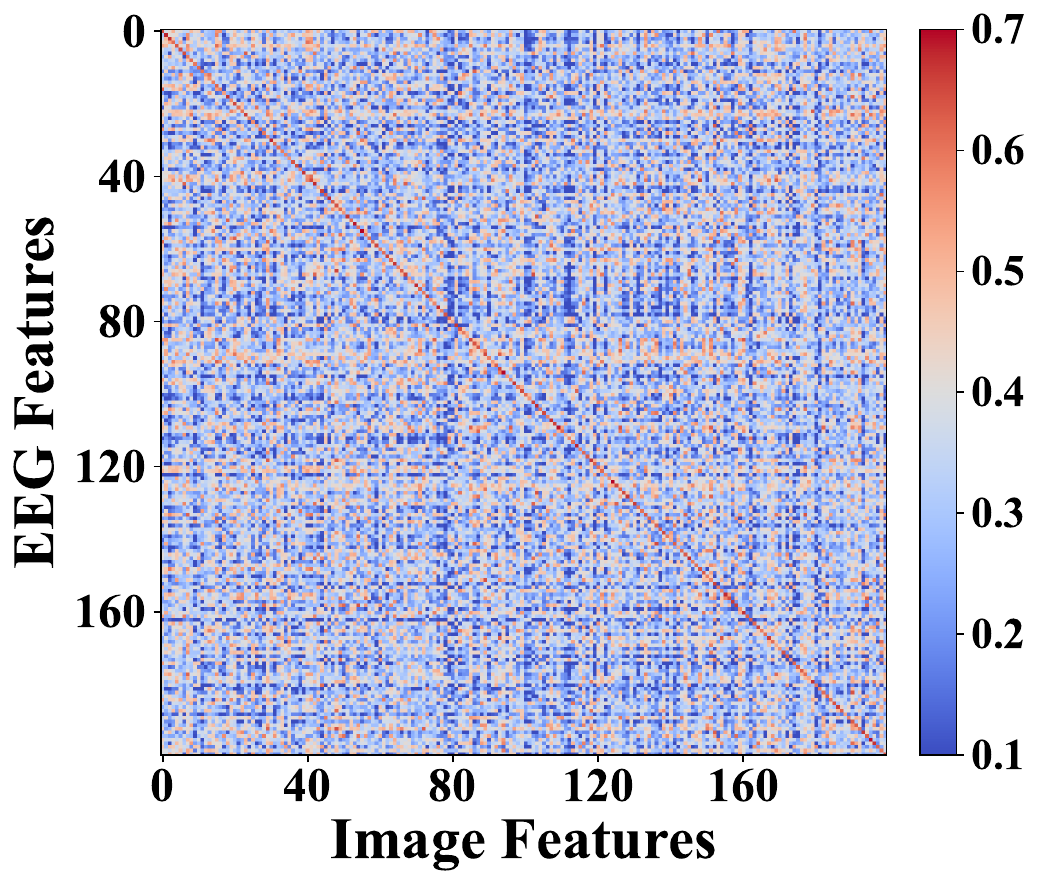}%
\label{fig4a}}
\hfil
\subfloat[]{\includegraphics[width=1.5in]{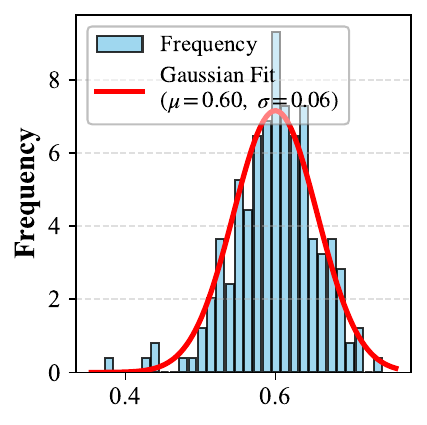}%
\label{fig4b}}
\caption{Similarity analysis of feature alignment. (a) Cross-modal similarity matrix between EEG and image features. The prominent diagonal indicates high similarity between corresponding sample pairs. (b) Statistical distribution of the diagonal elements in the similarity matrix, featuring a Gaussian fit with $\mu = 0.60$ and $\sigma = 0.06$.}
\label{fig4}
\end{figure}
\subsection{MI-Guided Dynamic Blur Regulation}
\label{sec:adaptive_blur} 
We introduce a self-adaptive optimization strategy designed to dynamically identify the optimal granularity of neuromimetic visual transformations. Our objective is to maximize the dependency between the EEG representations $F_{N}$ and the visual representation $F_{latent}$. Formally, we seek to maximize their Mutual Information (MI), denoted as $I(F_{N}; F_{latent})$, to ensure that the latent space captures the maximal shared semantic content across modalities. To align the $F_{N}$ and the $F_{latent}$ in the shared latent space, we optimize the symmetric contrastive loss:
\begin{equation}
\begin{split}
    \mathcal{L} = \frac{1}{2} \Big[ & \mathrm{CE}(sim(F_N, F_{{latent}}) / \tau,L_{batch}) \ + \\
    & \mathrm{CE}(sim(F_{{latent}}, F_N) / \tau,L_{batch}) \Big],
\end{split}
\label{eq:loss}
\end{equation}
where $\mathrm{CE}(\cdot)$ denotes the Cross-Entropy loss function, ${sim}(\cdot, \cdot)$ denotes the cosine similarity, $L_{batch}$ denotes the ground truth labels, and $\tau$ is a learnable temperature parameter. 

As direct MI estimation in high-dimensional space is intractable, we leverage the theoretical connection between contrastive learning and MI~\cite{oord2018representation}. The contrastive logit for a positive pair, denoted as $logits_{N}=sim(F_N, F_{{latent}})/\tau$, serves as an estimator of the log-density ratio associated with the pointwise mutual information. Accordingly, as visualized in the cross-modal similarity matrix in Fig.~\ref{fig4}(a), the diagonal elements of $logits_{N}$ serve as feedback signals indicating the alignment quality under the current blur parameters. To stabilize the feedback under mini-batch stochasticity and adaptively characterize alignment difficulty, we introduce a distribution-aware confidence mechanism. We apply momentum smoothing to instance-wise similarity scores to mitigate training variance. Let $s_{t}^{(i)}$ denote the $logits_{N}$ of the $i$-th sample at step $t$, and the smoothed similarity $\hat{s}_{t}^{(i)}$ is updated as:
\begin{equation}
    \hat{s}_{t}^{(i)} = \beta \cdot s_{t}^{(i)} + (1 - \beta) \cdot \hat{s}_{t-1}^{(i)},
    \label{eq:momentum_update}
\end{equation}
where $\beta$ is the momentum coefficient. Instead of using fixed thresholds, we dynamically construct a confidence interval based on the statistics of the current batch. As shown in Fig.~\ref{fig4}(b), similarity scores follow a Gaussian distribution during training dynamics~\cite{wang2025neuroclip}, we calculate the mean $\mu_s$ and standard deviation $\sigma_s$ of the smoothed similarities and define the acceptance bounds using the Z-score ($z_{\alpha/2}$) corresponding to a confidence level of $1-\alpha$:
\begin{equation}
    \begin{aligned}
        T_{upper} &= \mu_s + z_{\alpha/2} \cdot \sigma_s \\
        T_{lower} &= \mu_s - z_{\alpha/2} \cdot \sigma_s.
    \end{aligned}
    \label{eq:confidence_interval}
\end{equation}

Based on these dynamic bounds, we assign a new Gaussian blur configuration to each sample. Specifically, the blur configuration is jointly determined by the initial blur kernel size $k$ and the perturbation constant $c$ to update ${k}_{t}^{(i)}$ of the $i$-th sample at step $t$. In this way, the proposed feedback-driven blur selection implicitly promotes mutual information maximization between visual and EEG representations, enabling adaptive control of visual degradation based on alignment quality:
\begin{equation}
    {k}_{t}^{(i)} = 
    \begin{cases} 
    k - c, & \text{if } \hat s_{t}^{(i)} > T_{upper} \\
    k , & \text{if } T_{lower} < \hat s_{t}^{(i)} < T_{upper} \\
    k + c, & \text{if } \hat s_{t}^{(i)} < T_{lower}.  
    \end{cases}
    \label{eq:feedback_label}
\end{equation}

\begin{table*}[t]
\centering
\caption{TOP-1 and TOP-5 ACCURACY (\%) for 200-WAY ZERO-SHOT RETRIEVAL ON THINGS-EEG2.}
\label{table1}
\resizebox{\textwidth}{!}{%
\begin{tabular}{lcccccccccccccccccccccc}
\toprule
 & \multicolumn{2}{c}{Subject 1} & \multicolumn{2}{c}{Subject 2} & \multicolumn{2}{c}{Subject 3} & \multicolumn{2}{c}{Subject 4} & \multicolumn{2}{c}{Subject 5} & \multicolumn{2}{c}{Subject 6} & \multicolumn{2}{c}{Subject 7} & \multicolumn{2}{c}{Subject 8} & \multicolumn{2}{c}{Subject 9} & \multicolumn{2}{c}{Subject 10} & \multicolumn{2}{c}{Avg} \\
\cmidrule(lr){2-3} \cmidrule(lr){4-5} \cmidrule(lr){6-7} \cmidrule(lr){8-9} \cmidrule(lr){10-11} \cmidrule(lr){12-13} \cmidrule(lr){14-15} \cmidrule(lr){16-17} \cmidrule(lr){18-19} \cmidrule(lr){20-21} \cmidrule(lr){22-23}
Methods & Top-1 & Top-5 & Top-1 & Top-5 & Top-1 & Top-5 & Top-1 & Top-5 & Top-1 & Top-5 & Top-1 & Top-5 & Top-1 & Top-5 & Top-1 & Top-5 & Top-1 & Top-5 & Top-1 & Top-5 & Top-1 & Top-5 \\ \midrule
\multicolumn{23}{c}{\textbf{Intra-subject: train and test on one subject}} \\ \midrule
BraVL~\cite{du2023decoding}    & 6.1 & 17.9 & 4.9 & 14.9 & 5.6 & 17.4 & 5.0 & 15.1 & 4.0 & 13.4 & 6.0 & 18.2 & 6.5 & 20.4 & 8.8 & 23.7 & 4.3 & 14.0 & 7.0 & 19.7 & 5.8 & 17.5 \\
NICE-GA~\cite{songdecoding}     & 15.2 & 40.1 & 13.9 & 40.1 & 14.7 & 42.7 & 17.6 & 48.9 & 9.0 & 29.7 & 16.4 & 44.4 & 14.9 & 43.1 & 20.3 & 52.1 & 14.1 & 39.7 & 19.6 & 46.7 & 15.6 & 42.8 \\
ATM-S~\cite{li2024visual}  & 25.6 & 60.4 & 22.0 & 54.5 & 25.0 & 62.4 & 31.4 & 60.9 & 12.9 & 43.0 & 21.3 & 51.1 & 30.5 & 61.5 & 38.8 & 72.0 & 34.4 & 51.5 & 29.1 & 63.5 & 28.5 & 60.4 \\
UBP~\cite{wu2025bridging}    & 41.2 & 70.5 & 51.2 & 80.9 & 51.2 & 82.0 & 51.1 & 76.9 & 42.2 & 72.8 & 57.5 & 83.5 & 49.0 & 79.9 & 58.6 & 85.8 & 45.1 & 76.2 & 61.5 & 88.2 & 50.9 & 79.7 \\
ATS~\cite{wu2025shrinking}   & 53.0 & 79.0 & 62.0 & 87.5 & 61.5 & 89.0 & 57.0 & 86.5 & 55.0 & 84.0 & 68.0 & 90.5 & 53.0 & 84.0 & 66.5 & 91.0 & 58.5 & 86.0 & 67.5 & 89.0 & 60.2 & 86.7 \\
NeuroBridge~\cite{zhang2025neurobridge}  & 50.0 & 77.6 & 63.2 & 90.6 & 61.6 & 91.1 & 61.4 & 90.0 & 54.8 & 85.0 & 69.7 & 92.9 & 62.7 & 88.8 & 71.2 & 95.1 & 64.0 & 91.0 & 73.6 & 97.1 & 63.2 & 89.9 \\
\rowcolor{green!15} 
\textbf{BI-Cap} & \textbf{65.0} & \textbf{87.5} & \textbf{77.0} & \textbf{97.0} & \textbf{74.5} & \textbf{97.5} & \textbf{69.5} & \textbf{93.5} & \textbf{69.0} & \textbf{92.0} & \textbf{76.5} & \textbf{96.5} & \textbf{65.5} & \textbf{94.0} & \textbf{77.5} & \textbf{98.5} & \textbf{72.5} & \textbf{97.0} & \textbf{76.5} & \textbf{98.5} & \textbf{72.4} & \textbf{95.2} \\
\midrule
\multicolumn{23}{c}{\textbf{Inter-subject: leave one subject out for test}} \\ \midrule
BraVL~\cite{du2023decoding}         & 2.3 & 8.0 & 1.5 & 6.3 & 1.4 & 5.9 & 1.7 & 6.7 & 1.5 & 5.6 & 1.8 & 7.2 & 2.1 & 8.1 & 2.2 & 7.6 & 1.6 & 6.4 & 2.3 & 8.5 & 1.8 & 7.0 \\
NICE-GA~\cite{songdecoding}      & 5.9 & 21.4 & 6.4 & 22.7 & 5.5 & 20.1 & 6.1 & 21.0 & 4.7 & 19.5 & 6.2 & 22.5 & 5.9 & 19.1 & 7.3 & 25.3 & 4.8 & 18.3 & 6.2 & 26.3 & 5.9 & 21.6 \\
ATM-S~\cite{li2024visual}     & 10.5 & 26.8 & 7.1 & 24.8 & 11.9 & 33.8 & 14.7 & 39.4 & 7.0 & 23.9 & 11.1 & 35.8 & 16.1 & 43.5 & 15.0 & 40.3 & 4.9 & 22.7 & 20.5 & 46.5 & 11.8 & 33.7 \\
UBP~\cite{wu2025bridging}   & 11.5 & 29.7 & 15.5 & 40.0 & 9.8 & 27.0 & 13.0 & 32.3 & 8.8 & 33.8 & 11.7 & 31.0 & 10.2 & 23.8 & 12.2 & 32.2 & 15.5 & 40.5 & 16.0 & 43.5 & 12.4 & 33.4 \\
HA~\cite{rajabi2025human}   & 11.4 & 33.6 & 15.0 & 38.1 & 8.1 & 28.3 & 9.5 & 30.1 & 7.3 & 23.5 & 21.8 & 46.9 & 10.8 & 35.2 & 14.5 & 35.6 & 13.9 & 33.8 & 15.8 & 40.8 & 12.8 & 34.6 \\
ATS~\cite{wu2025shrinking}   & 17.0 & 37.5 & 21.5 & 49.5 & 7.5 & 20.0 & 13.5 & 41.5 & 7.0 & 27.0 & 14.5 & 37.0 & 10.0 & 33.5 & 13.0 & 30.5 & 9.5 & 28.5 & 26.0 & 52.5 & 14.0 & 35.8 \\
NeuroBridge~\cite{zhang2025neurobridge} & 23.2 & 52.4 & 21.2 & 49.3 & \textbf{13.2} & \textbf{36.5} & 17.0 & 45.3 & 14.5 & 37.7 & \textbf{25.0} & \textbf{55.0} & 15.3 & \textbf{45.1} & \textbf{20.1} & \textbf{44.9} & \textbf{13.7} & \textbf{36.5} & 27.2 & 56.3 & 19.0 & 45.9 \\
\rowcolor{green!15} 
\textbf{BI-Cap} & \textbf{24.5} & \textbf{54.5} & \textbf{28.5} & \textbf{62.0} & 11.5 & 29.5 & \textbf{20.0} & \textbf{48.5} & \textbf{15.5} & \textbf{41.5} & 15.0 & 44.5 & \textbf{20.5} & 41.5 & 13.5 & 37.0 & 13.0 & 36.0 & \textbf{31.5} & \textbf{65.0} & \textbf{19.4} & \textbf{46.0} \\
\midrule
\bottomrule
\end{tabular}
}
\end{table*}

\begin{table}[t]
\centering
\caption{TOP-1 and TOP-5 ACCURACY (\%) for 200-WAY ZERO-SHOT RETRIEVAL ON THINGS-MEG.}
\label{table2}
\resizebox{\columnwidth}{!}{%
\begin{tabular}{lcccccccccc}
\toprule
\multirow{2}{*}{Methods} & \multicolumn{2}{c}{Subject 1} & \multicolumn{2}{c}{Subject 2} & \multicolumn{2}{c}{Subject 3} & \multicolumn{2}{c}{Subject 4} & \multicolumn{2}{c}{Avg} \\
\cmidrule(lr){2-3} \cmidrule(lr){4-5} \cmidrule(lr){6-7} \cmidrule(lr){8-9} \cmidrule(lr){10-11}
 & Top-1 & Top-5 & Top-1 & Top-5 & Top-1 & Top-5 & Top-1 & Top-5 & Top-1 & Top-5 \\ \midrule
\multicolumn{11}{c}{\textbf{Intra-subject: train and test on one subject}} \\ \midrule
NICE-GA~\cite{songdecoding}       & 8.7 & 30.5 & 21.8 & 56.6 & 16.5 & 49.7 & 10.3 & 32.3 & 14.3 & 42.3 \\
HA~\cite{rajabi2025human}       & 15.5 & 39.6 & 36.8 & 75.2 & 25.6 & 55.0 & 14.7 & 38.6 & 23.2 & 52.1 \\
UBP~\cite{wu2025bridging}    & 15.0 & 38.0 & 46.0 & 80.5 & 27.3 & 59.0 & 18.5 & 43.5 & 26.7 & 55.2 \\
NeuroBridge~\cite{zhang2025neurobridge}     & 16.5 & 41.6 & 53.7 & 85.3 & 40.4 & 73.2 & 18.1 & 43.1 & 32.2 & 60.8 \\
ATS~\cite{wu2025shrinking}         & 18.0 & 43.0 & 56.0 & 89.0 & 36.0 & 67.0 & 19.5 & 50.0 & 32.4 & 62.3 \\
\rowcolor{green!15} 
\textbf{BI-Cap} & \textbf{19.0} & \textbf{45.5} & \textbf{70.0} & \textbf{94.0} & \textbf{48.5} & \textbf{82.0} & \textbf{24.0} & \textbf{54.5} & \textbf{40.4} & \textbf{69.0} \\ 
\midrule
\multicolumn{11}{c}{\textbf{Inter-subject: leave one subject out for test}} \\ \midrule
UBP~\cite{wu2025bridging}          & 2.0 & 5.7 & 1.5 & 17.2 & 2.7 & 10.5 & 2.5 & 8.0 & 2.2 & 10.4 \\
ATS~\cite{wu2025shrinking}          & 2.0 & 6.0 & 6.0 & 14.0 & 4.5 & 15.0 & 1.0 & 9.5 & 3.4 & 11.2 \\
NeuroBridge~\cite{zhang2025neurobridge}     & \textbf{4.3} & \textbf{13.1} & 3.6 & 15.6 & 3.0 & 11.2 & \textbf{2.5} & \textbf{11.3} & 3.4 & 12.8 \\
\rowcolor{green!15} 
\textbf{BI-Cap} & 2.5 & 6.0 & \textbf{6.0} & \textbf{22.0} & \textbf{5.0} & \textbf{16.5} & 2.0 & 9.5 & \textbf{3.9} & \textbf{13.5} \\ \midrule
\bottomrule
\end{tabular}
}
\end{table}

\section{Experimental Setup}
\subsection{Implementation Details}
Our experiments are carried out using the PyTorch framework on a Linux server platform equipped with an NVIDIA Tesla V100. All models were trained for 150 epochs using the AdamW optimizer with weight decay. For the THINGS-EEG dataset, we set the batch size to 32 and the learning rate to $1 \times 10^{-4}$, with the initial Gaussian blur kernel size $k$ configured to 75. Conversely, for the THINGS-MEG dataset, we utilized a larger batch size of 1024 and a learning rate of $3 \times 10^{-4}$, setting the Gaussian blur kernel size to 51. Regarding the hyperparameters of our proposed method, the temperature parameter $\tau$ for the contrastive loss was set to 0.07. The perturbation constant $c$ was set to 6, and the standard deviation $\sigma$ for the Gaussian noise was set to 10. Finally, the sampling scale factors were set to $s_{low} = 1/2$ and $s_{mos} = 1/16$.
\subsection{Image and Brain Encoder}
We leverage pre-trained weights from CLIP~\cite{radford2021learning} and OpenCLIP~\cite{cherti2023reproducible} to extract visual embeddings, employing a diverse range of backbones including ResNet-50, ResNet-101, ViT-B/16, ViT-B/32, ViT-L/14, ViT-H/14, ViT-g/14, and ViT-bigG/14. Unless otherwise specified, we utilize ResNet-50 as the default backbone for the THINGS-EEG2 dataset and ViT-B/16 for the THINGS-MEG dataset. For brain encoder, we adopt the Shared Temporal Attention Encoder (STAE) proposed by~\cite{wu2025shrinking}. This architecture incorporates a shared temporal attention mechanism designed to mitigate the temporal aliasing effects inherent in the RSVP paradigm. Specifically, the STAE aims to automatically focus on the time steps most relevant to visual decoding by assigning greater attention weights to the most informative temporal segments of the neural signals.
\subsection{Datasets and Evaluation Metrics}
\label{sec:datasets}
We evaluate our method on two large-scale neuroimaging datasets: THINGS-EEG2~\cite{gifford2023large} and THINGS-MEG~\cite{hebart2023things}, both of which are based on the THINGS object concept database. Following the zero-shot protocol established in~\cite{wu2025bridging}, the training and test concepts are strictly mutually exclusive to ensure an unbiased assessment of semantic generalization.
\subsubsection{THINGS-EEG2~\cite{gifford2023large}}
This dataset consists of 64-channel EEG recordings from 10 subjects collected using an EASYCAP system under the RSVP paradigm. The training set comprises 1,654 object concepts, with 10 images per concept and each image repeated four times per subject. The test set contains 200 unseen concepts, each represented by a single image repeated 80 times to facilitate robust zero-shot evaluation. For EEG preprocessing, raw signals were band-pass filtered between 0.1 and 100 Hz and subsequently downsampled to 250 Hz. To improve the Signal-to-Noise Ratio (SNR), trial repetitions are averaged, yielding a final dataset of 16,540 training samples and 200 test samples per subject. Especially, we implement different channel selection strategies for experimental settings of intra and inter subject. For intra-subject, where neural patterns are consistent, we selected a subset of 17 posterior channels located over the occipito-parietal cortex (P7, P5, P3, P1, Pz, P2, P4, P6, P8, PO7, PO3, POz, PO4, PO8, O1, Oz, O2) to concentrate on visual cortex activity. To address the significant heterogeneity in the inter-subject generalization tasks, we expanded our input to include the full montage of 63 channels, thereby leveraging global brain activity to learn more robust subject-invariant features. 
\subsubsection{THINGS-MEG~\cite{hebart2023things}}
This dataset contains high-density recordings from 4 subjects using a 271-channel MEG system. The training set includes 1,854 concepts featuring 12 unique images per concept, with each image presented a single time. The test set comprises 200 unseen concepts, each represented by one image that was repeated 12 times to ensure robust evaluation. MEG data were band-pass filtered between 0.1 and 100 Hz, baseline corrected, and downsampled to 200 Hz. To maximize the SNR, multiple repetitions of the same stimulus are averaged into a single representative trial. We consistently utilized the full array of 271 sensors across all experimental settings. This holistic approach ensures that the fine-grained spatiotemporal details and the complex magnetic field distributions inherent in high-density MEG recordings are fully preserved for cross-modal alignment.

To quantitatively evaluate the efficacy of our proposed BI-Cap framework, we employ three key metrics: Top-$n$ Accuracy, mean Average Precision (mAP), and Similarity Score. Collectively, these metrics provide a holistic assessment of both ranking quality and intrinsic semantic alignment.

\section{Results}
\subsection{Comparison Performance}
To validate the effectiveness of our proposed approach, we benchmark it against a comprehensive set of recent neural decoding methods, including BraVL~\cite{du2023decoding}, NICE~\cite{songdecoding}, ATM-S~\cite{li2024visual}, UBP~\cite{wu2025bridging}, HA~\cite{rajabi2025human}, ATS~\cite{wu2025shrinking}, and the previous state-of-the-art method NeuroBridge~\cite{zhang2025neurobridge}. The methodological contributions of these works have been discussed in detail in Section~\ref{section2}. For a fair and rigorous comparison, all baseline results are directly taken from the best reported performances in the original papers and we maintain complete consistency in data preprocessing pipelines across all experiments.

Table~\ref{table1} and Table~\ref{table2} report the quantitative performance of BI-Cap against competing baselines on zero-shot brain-to-image retrieval tasks for THINGS-EEG2 and THINGS-MEG datasets. Our method consistently outperforms previous state-of-the-art approaches under intra-subject and inter-subject settings, demonstrating superior feature alignment and generalization capabilities.
Notably, our method achieves a Top-1 accuracy of 72.4\% and a Top-5 accuracy of 95.2\% on the THINGS-EEG2, yielding the best performance across all subjects and surpassing the previous state-of-the-art Top-1 accuracy by a margin of 9.2\%.
Similarly, on the THINGS-MEG, our model attains a Top-1 accuracy of 40.4\% and a Top-5 accuracy of 69.0\%, outperforming the prior best Top-1 result by 8.0\% across all subjects. These results suggest that our biomimetic transformation and evidence-driven fusion effectively capture the core semantic components.
\begin{table}[t]
\centering
\caption{ABLATION STUDY of PROGRESSIVE COMPONENTS ON THINGS-EEG2 UNDER the INTRA-SUBJECT SETTING. WE INCREMENTALLY ADD DYNAMIC FOVEATION (DYN.), GAUSSIAN NOISE (NOISE.), LOW-RESOLUTION (RES.), MOSAIC (MOS.), and EVIDENCE LEARNING (EL).}
\label{table3}
\begin{tabular}{ccccccc}
\toprule
\multicolumn{5}{c}{Components} & \multicolumn{2}{c}{Avg} \\
\cmidrule(lr){1-5} \cmidrule(lr){6-7}
Dyn. & Noise. & Res. & Mos. & EL & Top-1 (\%) & Top-5 (\%) \\ \midrule
 & & & &  &  48.9 & 81.8 \\
\checkmark & & & &  & 58.8 & 87.1 \\
\checkmark & \checkmark & & &  & 63.7 & 90.2 \\
\checkmark & \checkmark & \checkmark & &  & 66.4 & 92.7 \\
\checkmark & \checkmark & \checkmark & \checkmark & & 70.4 & 94.1 \\
\rowcolor{green!15}
\checkmark & \checkmark & \checkmark & \checkmark &  \checkmark  & \textbf{72.4} & \textbf{95.2} \\
\bottomrule
\end{tabular}%
\end{table}
\begin{table}[htbp]
  \centering
  \caption{ABLATION STUDY of PROGRESSIVE COMPONENTS ON THINGS-MEG UNDER the INTER-SUBJECT SETTING. WE INCREMENTALLY ADD DYNAMIC FOVEATION (DYN.), GAUSSIAN NOISE (NOISE.), LOW-RESOLUTION (RES.), MOSAIC (MOS.), and EVIDENCE LEARNING (EL).}
  \label{table4}
  \begin{tabular}{ccccccc}
    \toprule
    \multicolumn{5}{c}{Components} & \multicolumn{2}{c}{Avg} \\
    \cmidrule(lr){1-5} \cmidrule(lr){6-7}
    Dyn. & Noise. & Res. & Mos. & EL & Top-1 (\%) & Top-5 (\%) \\
    \midrule
    \checkmark &            &            &            &            & 2.5          & 11.4          \\
    \checkmark & \checkmark &            &            &            & 2.6          & 11.5          \\
    \checkmark & \checkmark & \checkmark &            &            & 2.4          & 10.0          \\
    \checkmark & \checkmark & \checkmark & \checkmark &            & 3.6          & 12.1          \\
    \rowcolor{green!15}
    \checkmark & \checkmark & \checkmark & \checkmark & \checkmark & \textbf{3.9} & \textbf{13.5} \\
    \bottomrule
  \end{tabular}
\end{table}

Notably, BI-Cap exhibits competitive robustness in the more challenging inter-subject scenarios, where significant neural heterogeneity leads to severe performance decline for existing models. On the THINGS-EEG2, BI-Cap maintains a leading average Top-1 accuracy of 19.4\%, and on the THINGS-MEG, BI-Cap also achieves the highest average performance.

\begin{figure}[t]
\centering
\subfloat[]{\includegraphics[width=1.5in]{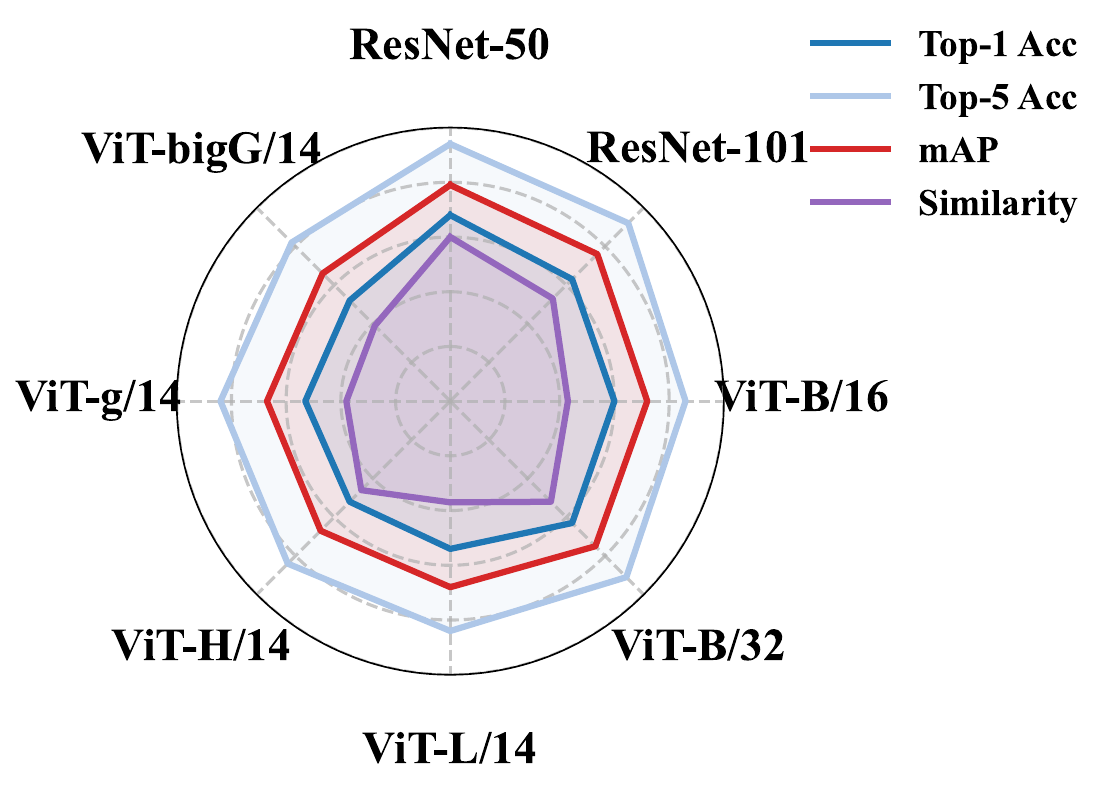}%
\label{fig:radar}}
\hfil
\subfloat[]{\includegraphics[width=1.5in]{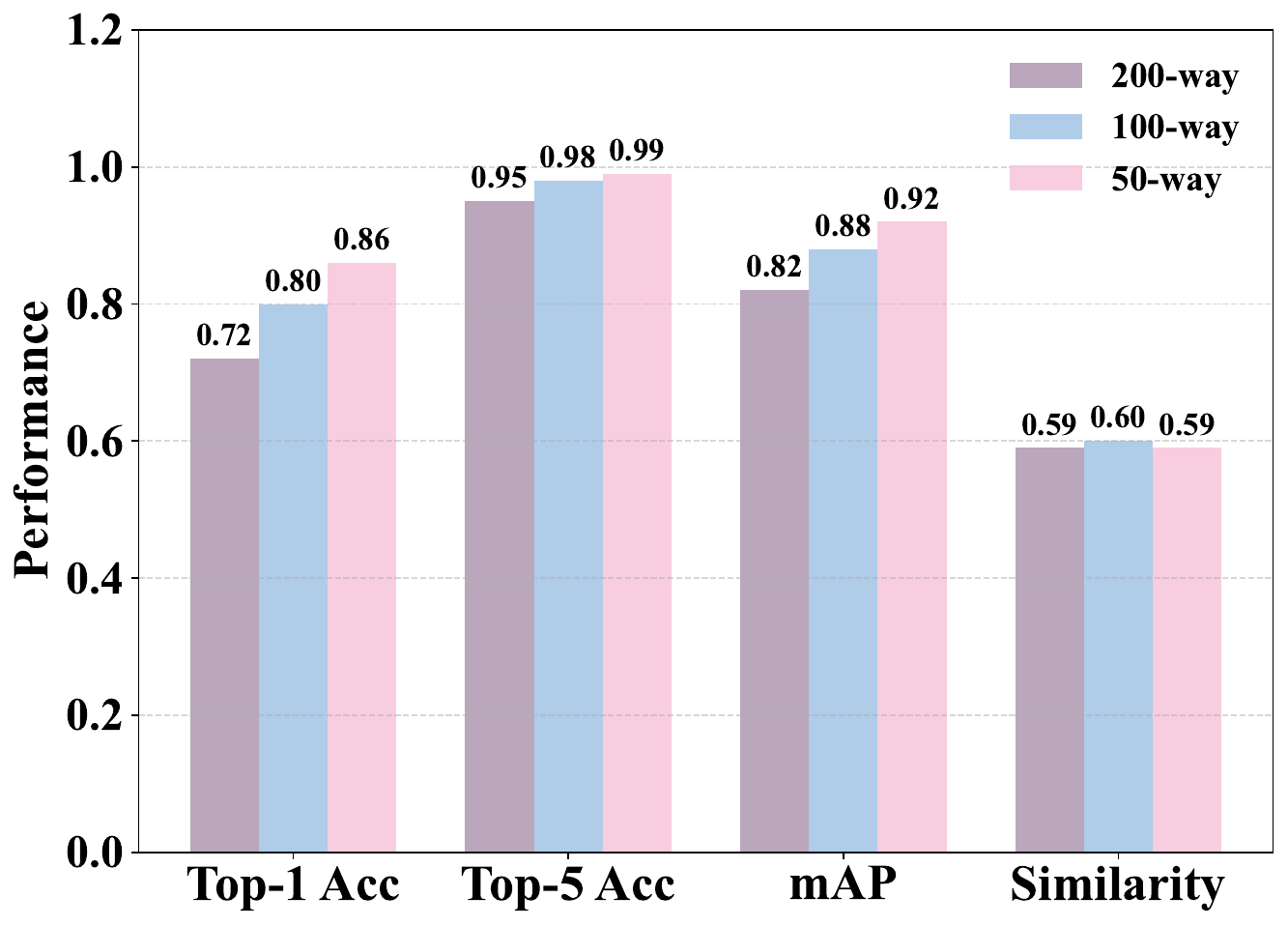}%
\label{fig:line}}
\vspace{1pt}
\subfloat[]{\includegraphics[width=1.5in]{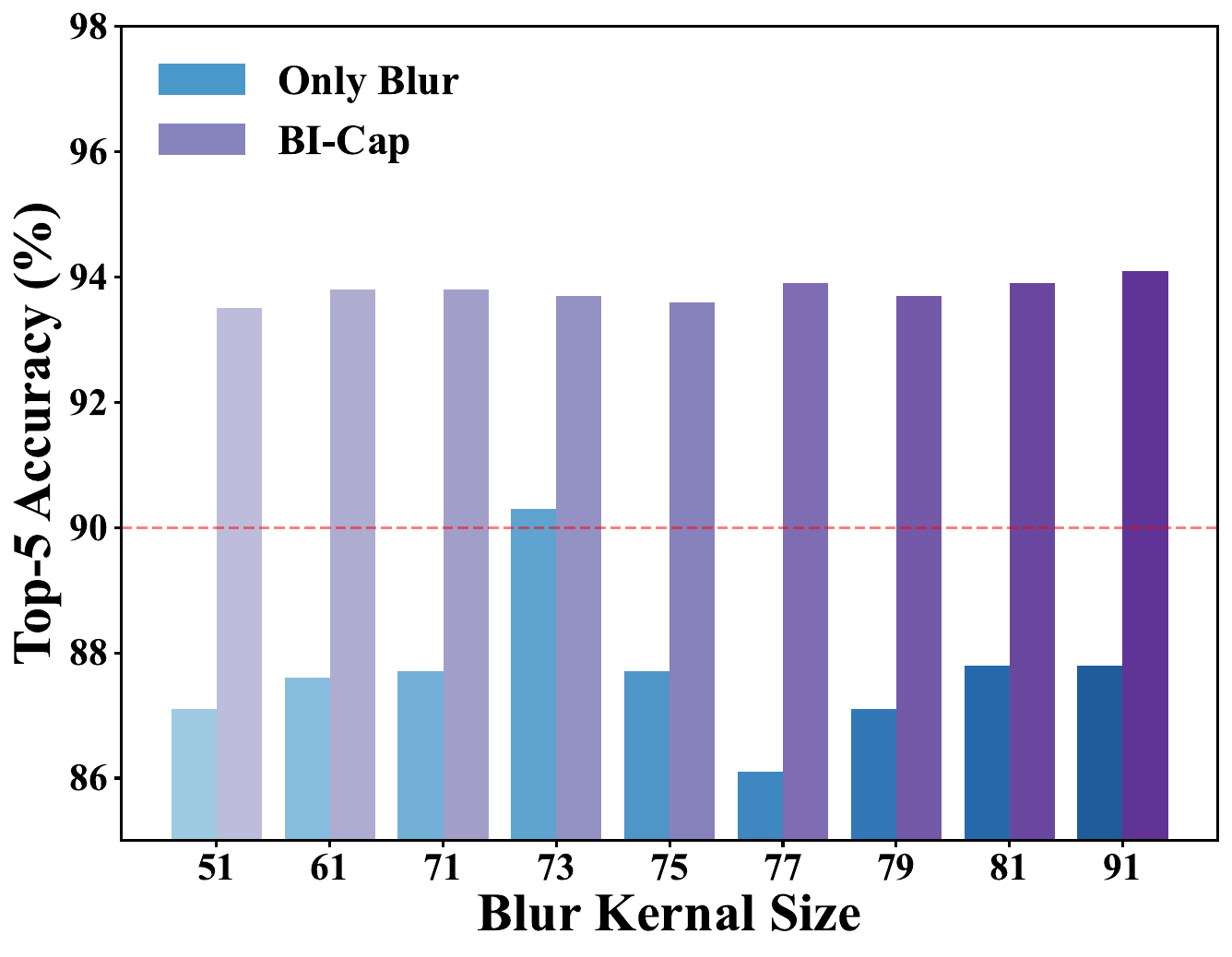}%
\label{fig:c_image}}
\hfil
\subfloat[]{\includegraphics[width=1.5in]{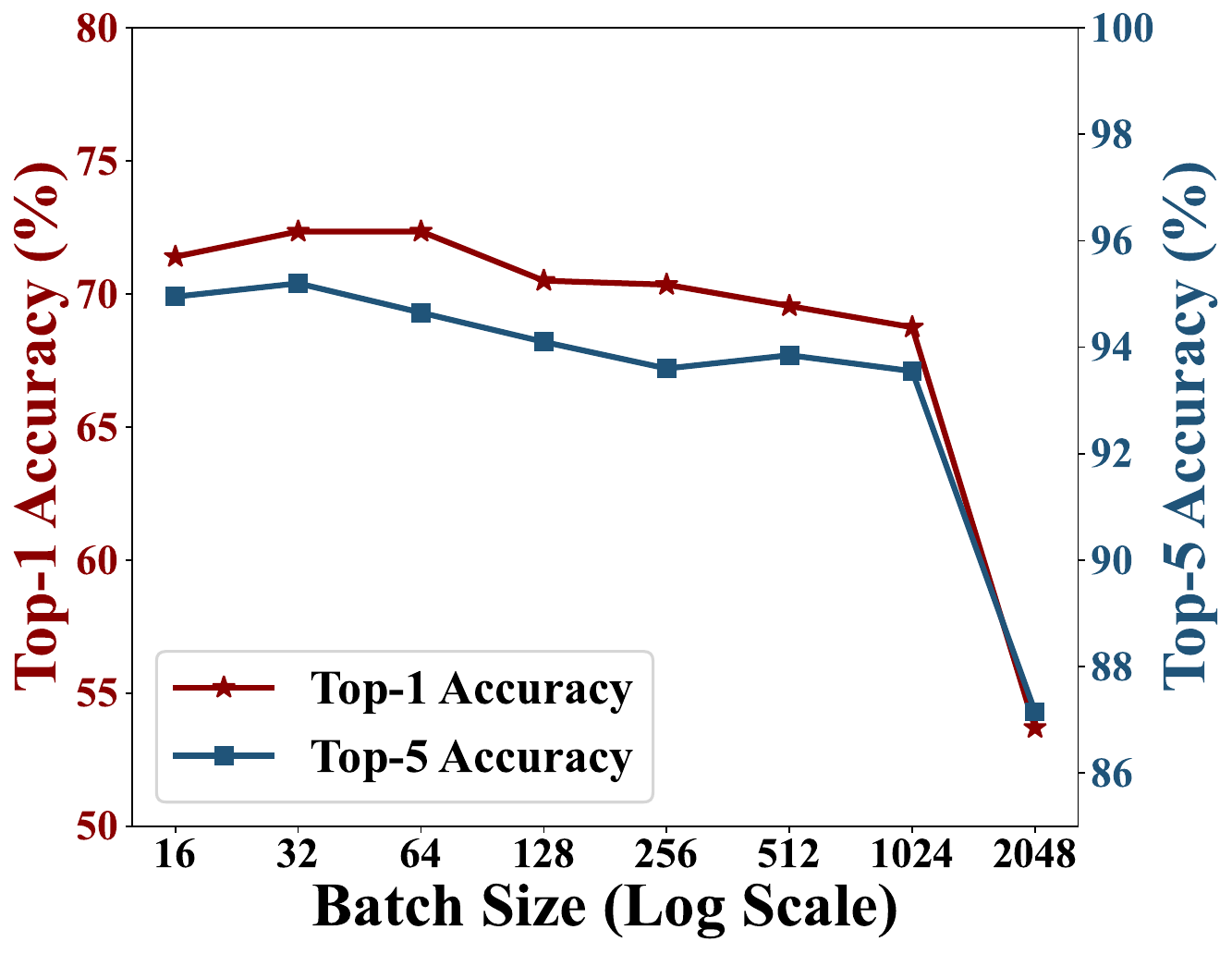}%
\label{fig:d_image}}
\caption{All results presented in the figure are averaged across 10 subjects in the intra-subject setting on THINGS-EEG2. (a) Comparison of different image encoders. (b) Performance across different retrieval set sizes. (c) Sensitivity analysis of blur kernel size. (d) The impact of batch size on retrieval accuracy.}
\label{fig5}
\end{figure}

\subsection{Ablation Performance}
\subsubsection{Effectiveness of Biologically Plausible Transformations and Evidence Learning}
To evaluate the effectiveness of four biologically transformations and evidence learning proposed, we conducted a progressive ablation study on both THINGS-EEG2 and THINGS-MEG. The quantitative results are detailed in Table~\ref{table3} for the intra-subject setting and Table~\ref{table4} for the inter-subject task, respectively. Dynamic Foveation, Gaussian Noise, Low Resolution, and Mosaic transformations are sequentially incorporated into the training pipeline. 

In the intra-subject setting of THINGS-EEG2, performance improves monotonically with each successive addition, confirming that our paradigm effectively strengthens the cross-modal alignment between neural and visual representations. For the more challenging inter-subject task of THINGS-MEG, although the absolute improvement is subject to significant intrinsic heterogeneity in underlying physiology, the overall trend remains consistent, with the combined transformations yielding superior results compared to the baseline.

We further validated the effectiveness of EL as an uncertainty-aware mechanism. Across both datasets and experimental settings, the integration of EL consistently provides the final performance boost. Specifically, in the intra-subject setting of THINGS-EEG2, adding EL boosts the Top-1 accuracy from 70.4\% to 72.4\%; in the inter-subject setting of THINGS-MEG, it further elevates the Top-1 accuracy to its peak at 3.9\%. These results indicate that explicitly modeling uncertainty acts as a critical regularizer, enabling the network to discern reliable features and achieve robust alignment even under the high-variance conditions of inter-subject decoding.
\subsubsection{Effect of Blur Strategies}
As shown in Table~\ref{table5},  the proposed dynamic strategy achieves superior performance across all metrics, surpassing the static strategy by a clear margin. This result indicates that dynamic central concavity improves performance by mimicking the spatial attention mechanism of the HVS, emphasizing salient regions while attenuating peripheral information, which facilitates the extraction of more biologically plausible features than static.
\subsubsection{Effect of Image Encoders}
We conduct an ablation study with a batch size of 1024 across eight image encoders to analyze the impact of visual feature structure on EEG decoding. As shown in Fig.~\ref{fig5}(a) and Table~\ref{table6}, scaling up model size does not improve decoding performance. Instead, the CNN-based ResNet-50 achieves the best results, consistently outperforming Transformer-based models across all evaluation metrics. This observation suggests that the hierarchical inductive bias of CNNs is more compatible with EEG decoding, whereas the globally attended feature representations of ViT models may be less suitable for effective EEG–visual alignment.

\begin{figure*}[!t]
\centering
\subfloat[]{\includegraphics[width=3in]{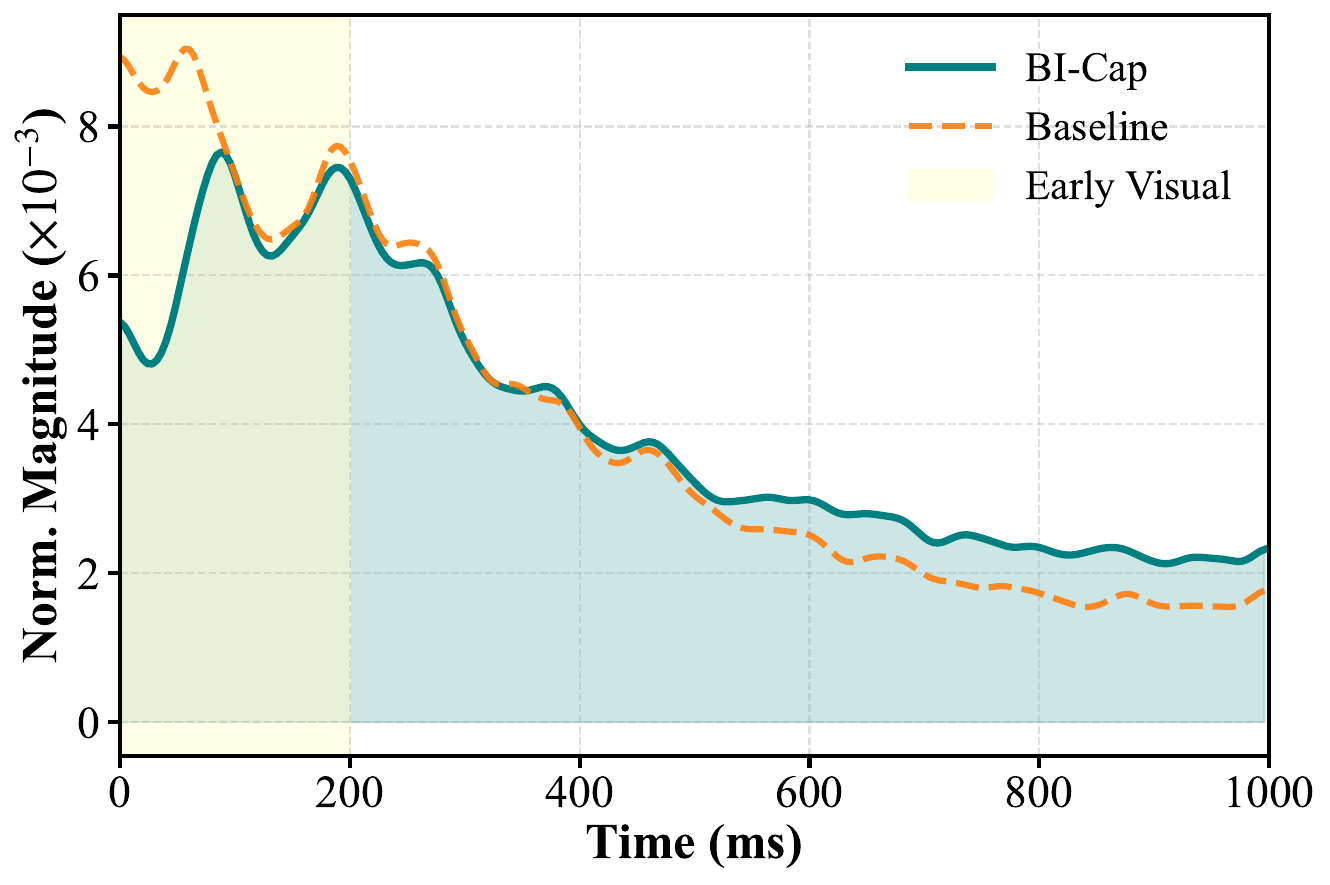}%
\label{fig4:temporal}}
\hfil
\subfloat[]{\includegraphics[width=3in]{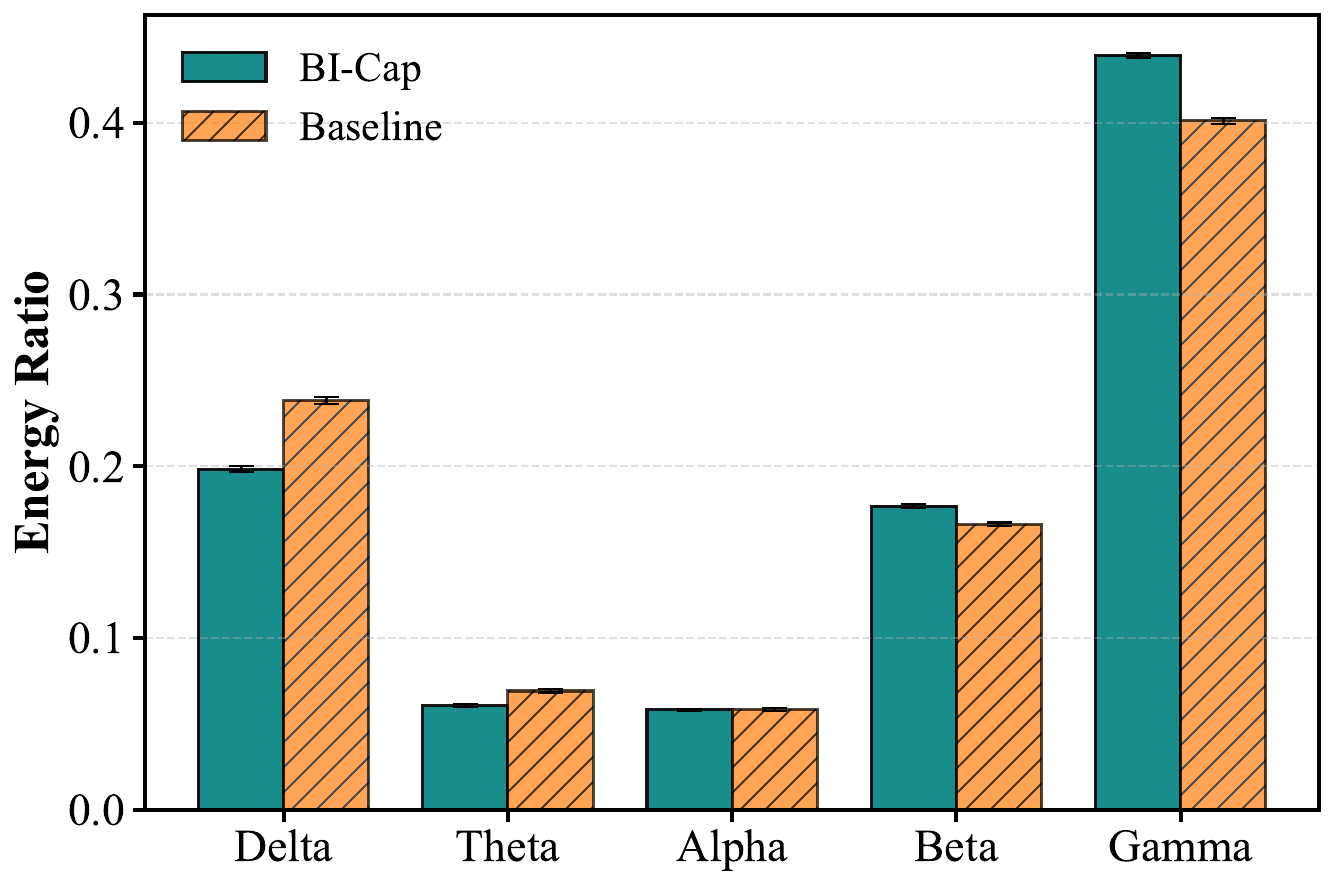}%
\label{fig4:spectral}}
\caption{Comparison of temporal and spectral gradient analysis on THINGS-EEG2 for subject 4 with baseline (ATS). (a) Temporal gradient distribution. (b) Spectral gradient distribution.}
\label{fig6}
\end{figure*}

\subsection{Evaluation under Different Retrieval Sizes}
o evaluate the robustness of the model under varying retrieval difficulties, we compared the decoding performance across discrete candidate pool sizes of 200, 100, and 50, while maintaining a fixed ResNet-50 backbone. As illustrated in Fig.~\ref{fig5}(b), reducing the retrieval space leads to consistent improvements of all metrics. In the 50-way setting, Top-5 accuracy approaches saturation (99\%), demonstrating strong discriminative capability.
Notably, the similarity score remains remarkably stable regardless of the number of candidates. This robust stability serves as a compelling indicator that the intrinsic cross-modal alignment between EEG representations and visual semantics is fundamentally preserved, rendering the learned features agnostic to external computational constraints.
\subsection{Hyperparameter Sensitivity Analysis}
\subsubsection{Sensitivity Analysis of Blur Kernal Size}
To evaluate robustness, we analyze retrieval accuracy under varying blur kernel sizes $k$. As shown in Fig.~\ref{fig5}(c), only blur exhibit strong sensitivity, since approaches such as UBP~\cite{wu2025bridging} rely on manually predefined kernel sizes and suffer sharp performance degradation once deviating from the optimum. In contrast, our method maintains a flat and stable performance curve, indicating that the proposed evidence-driven multi-neuromimetic framework effectively alleviates sensitivity to manual hyper-parameter choices by adaptively down-weighting unreliable branches.
\subsubsection{Sensitivity Analysis of Batch Size}
We investigated the impact of batch size on the retrieval performance, as this parameter serves as a critical determinant in contrastive learning by modulating the density of the negative sample pool available per stochastic iteration. As shown in Fig.~\ref{fig5}(d), our model demonstrates strong robustness across batch sizes ranging from 16 to 1024, with peak performance 32. This suggests that effective cross-modal alignment can be achieved without large-batch training. Performance drops only at an extreme batch size of 2048, likely due to optimization and generalization issues commonly observed in large-batch regimes.
\begin{table}[t]
\centering
\caption{COMPARISON of DIFFERENT BLUR STRATEGIES ON THINGS-EEG2. WE REPORT AVERAGE TOP-1/TOP-5 ACCURACY, MAP, and SIMILARITY.}
\label{table5}
\begin{tabular}{lcccc}
\toprule
\multirow{2}{*}{Blur Strategy} & \multicolumn{4}{c}{Avg} \\
\cmidrule(lr){2-5}
& Top-1 (\%) & Top-5 (\%) & mAP (\%) & Similarity \\ \midrule
Static & 70.5 & 94.8 & 80.8 & 0.585 \\
\textbf{Dynamic (Ours)} & \textbf{72.4} & \textbf{95.2} & \textbf{82.1} & \textbf{0.598} \\
\bottomrule
\end{tabular}%
\end{table}

\subsection{Biological Interpretation} 
To verify the biological consistency of our model, we visualized the input gradients to identify the specific temporal-spectral features that exert the most significant influence on the retrieval task. We employed the absolute gradient magnitudes as saliency maps to quantify and compare feature importance. To ensure a rigorous and controlled assessment, ATS~\cite{wu2025shrinking} was selected as the primary baseline because it utilizes a brain encoder architecture identical to our own.

For temporal analysis, as visualized in Fig.~\ref{fig6}(a), our method aligns with human visual attention dynamics. Unlike the Baseline, which exhibits distinctively high amplitudes in the artifact-dominant phase (0--100~ms), our method actively suppresses these early noise signals. Subsequently, it peaks at 100--200~ms, matching the P1/N1 latency for early attentional gain control~\cite{hillyard1998event}. Crucially, while the Baseline decays rapidly in the late period ($>200$~ms), our method sustains significantly higher gradient magnitudes. This persistent activity indicates prolonged semantic processing rather than the transient sensory registration observed in the baseline.

For spectral analysis, as visualized in Fig.~\ref{fig6}(b), we applied Welch's method to estimate the power spectral density of the gradient signals. We aggregated the spectral energy into standard frequency bands, Delta (0--4~Hz), Theta (4--8~Hz), Alpha (8--12~Hz), Beta (12--30~Hz), and Gamma (30--100~Hz), to determine the frequency specific contribution to the model's decision making process. The results reveal a distinct spectral shift in our method compared to baseline. Crucially, our model exhibits a dominant reliance on Gamma band activity. This is consistent with prior findings showing that induced gamma-band activity reflects higher-level visual integration and object representation~\cite{tallon1999oscillatory}. The baseline focuses predominantly on Delta, whereas our method emphasizes Beta and Gamma. This aligns with prior research linking these rhythms to feedback and feedforward influences in visual cortical areas~\cite{michalareas2016alpha}.
\begin{table*}
\centering
\caption{Performance comparison of different Image Encoder backbones. Experiments are conducted on the THINGS-EEG2 dataset using the intra-subject setting with 200-way zero-shot retrieval.}
\label{table6}
\resizebox{0.8\textwidth}{!}{%
\begin{tabular}{lcccccccc}
\toprule
\textbf{\makecell{Image \\ Encoder}} & ResNet-50 & ResNet-101 & ViT-B/16 & ViT-B/32 & ViT-L/14 & ViT-H/14 & ViT-g/14 & ViT-bigG/14 \\ \midrule
mAP (\%) & \textbf{79.0} & 75.5 & 71.7 & 74.8 & 67.6 & 66.6 & 66.7 & 65.6\\ 
Similarity & \textbf{0.598} & 0.525 & 0.448 & 0.516 & 0.369 & 0.459 & 0.381 & 0.388\\ 
Parameters & 5.6M & 4.9M & 3.4M & 3.1M & 5.8M & 5.6M & 6.1M & 7.2M\\ 
\bottomrule
\end{tabular}%
}
\end{table*}

\begin{figure*}
\centerline{\includegraphics[width=0.8\textwidth]{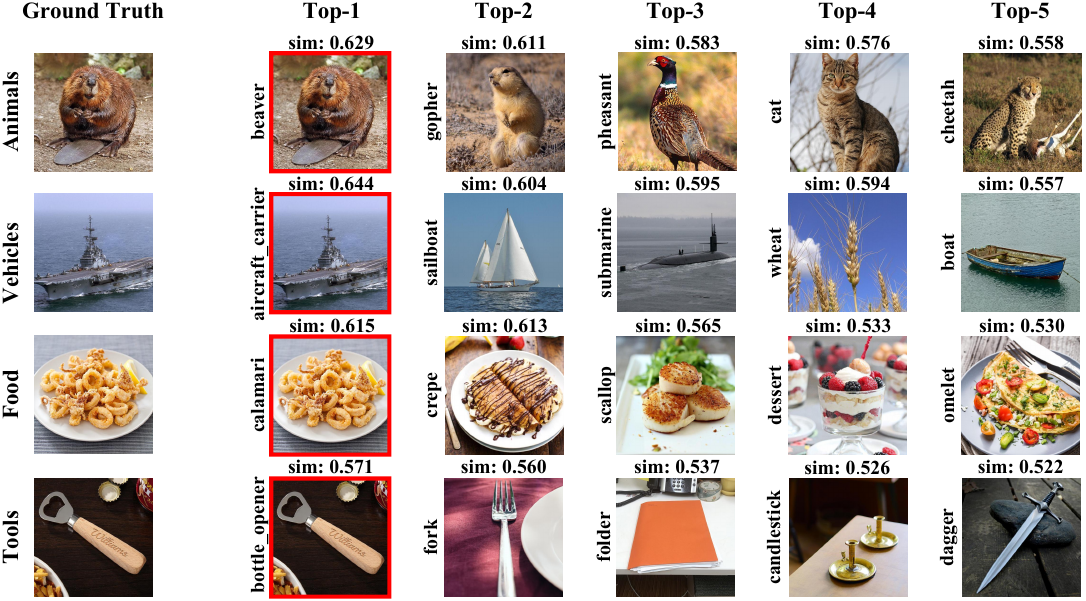}}
\caption{Top-5 retrieval visualization and semantic analysis for Subject 8. The first column displays ground-truth images across four semantic categories including Animals, Vehicles, Food, and Tools. Each retrieved image is presented with its predicted label and the corresponding similarity score (sim).} \label{fig7}
\end{figure*}

\subsection{Semantic Concept Retrieval}
In Fig.~\ref{fig7}, we visualize the concept-level retrieval results. We employ Representational Similarity Analysis (RSA)~\cite{cichy2020eeg} to examine the cross-modal alignment. Taking Subject 8 as an example, we present successful retrieval examples across diverse concepts, including \textit{Animals}, \textit{Vehicles}, \textit{Food}, and \textit{Tools}. 
Our model consistently achieves precise Top-1 retrieval, successfully mapping decoded EEG features back to their corresponding ground-truth images. Beyond Top-1 accuracy, the Top-5 retrieved images exhibit significant semantic proximity to the target, demonstrating effective preservation of concept-level semantics in decoded EEG features. Even when semantically unrelated images appear, they share similar low-level attributes, such as color and pattern, with the ground truth.

\section{Conclusion and Limitation}
In this work, we present BI-Cap, a neuromimetic perceptual simulation paradigm to bridge the systematic and stochastic gaps in neural visual decoding. By integrating biologically plausible static and dynamic visual transformations, we significantly enhance cross-modal alignment. Furthermore, to address neural dynamicity and heterogeneity, we introduce evidence-driven latent space representation. Our method achieves state-of-the-art performance in zero-shot brain-to-image retrieval. This work provides a principled computational perspective on biologically inspired visual decoding and offers a promising foundation for robust non-invasive BCIs systems.

While our BI-Cap framework demonstrates state-of-the-art performance in zero-shot brain-to-image retrieval, there are several limitations inherent to the current study that outline promising directions for future research. Although retrieval is a critical step in verifying semantic alignment, it is constrained by the size and diversity of the pre-defined gallery. To overcome this, our future work aims to transition from retrieval to visual reconstruction. By leveraging the evidence-driven latent representations learned by BI-Cap, we plan to condition generative models to reconstruct pixel-level images directly from neural signals. Despite the improvements achieved by BI-Cap over baseline methods in cross-subject scenarios, a significant performance gap remains between intra-subject and inter-subject decoding. The absolute accuracy in cross-subject retrieval is still constrained by the inherent inter-individual variability of EEG signals. These discrepancies arise from fundamental factors such as neuroanatomical differences, variations in electrode impedance, and distinct cognitive processing styles among participants. Consequently, achieving a truly universal neural decoder that generalizes seamlessly across diverse populations remains an ongoing challenge.

\bibliography{example_paper}
\bibliographystyle{IEEEtran}
\end{document}